\documentclass[switch]{article}
\usepackage{preprint}
\usepackage{algorithm}
\usepackage{algorithmic}
\usepackage{amsmath, amsthm, amssymb, amsfonts}

\usepackage[numbers,square]{natbib}
\usepackage[utf8]{inputenc}
\usepackage[T1]{fontenc}
\usepackage{xcolor}
\usepackage[colorlinks = true,
            linkcolor = purple,
            urlcolor  = blue,
            citecolor = cyan,
            anchorcolor = black]{hyperref}
\usepackage{booktabs}
\usepackage{nicefrac}
\usepackage{microtype}
\usepackage{lineno}
\usepackage{float}
\usepackage{lipsum}
\usepackage{newfloat}
\DeclareFloatingEnvironment[name={Supplementary Figure}]{suppfigure}
\usepackage{sidecap}
\sidecaptionvpos{figure}{c}

\usepackage{titlesec}
\titlespacing\section{0pt}{12pt plus 3pt minus 3pt}{1pt plus 1pt minus 1pt}
\titlespacing\subsection{0pt}{10pt plus 3pt minus 3pt}{1pt plus 1pt minus 1pt}
\titlespacing\subsubsection{0pt}{8pt plus 3pt minus 3pt}{1pt plus 1pt minus 1pt}

\usepackage{eso-pic}
\usepackage{tikz}
\usepackage{xcolor}
\PassOptionsToPackage{colorlinks=true,linkcolor=gray,urlcolor=gray}{hyperref}

\usepackage{titling}
\usepackage{footmisc}
\usepackage{multirow}
\usepackage{diagbox}
\newcommand{\Author}[2]{\textbf{#1}\textsuperscript{#2}}

\title{MambaBack: Bridging Local Features and Global Contexts in Whole Slide Image Analysis}

\author{
  \Author{Sicheng Chen}{1}\and
  \Author{Chad Wong}{1}\and
  \Author{Tianyi Zhang}{2} \and
  \Author{Enhui Chai}{3} \and
  \Author{Zeyu Liu}{3} \and
  \Author{Fei Xia}{1}
}

\date{
  \textsuperscript{1}Nhu Department of Electrical Engineering and Computer Science, University of California, Irvine \\
  \textsuperscript{2}Department of Electrical \& Computer Engineering, National University of Singapore \\
  \textsuperscript{3}PuzzleLogic Pte Ltd, Singapore 229594, Singapore \\
  [1em]
  \footnotesize \textbf{Corresponding author:} Zeyu Liu\texttt{<zeyuliu@puzzlelogic.com>}, Fei Xia\texttt{<fei.xia@uci.edu>} \\
}

\begin{document}
\maketitle
\begin{abstract}
Whole Slide Image (WSI) analysis is pivotal in computational pathology, enabling cancer diagnosis by integrating morphological and architectural cues across magnifications. Multiple Instance Learning (MIL) serves as the standard framework for WSI analysis. Recently, Mamba has become a promising backbone for MIL, overtaking Transformers due to its efficiency and global context modeling capabilities originating from Natural Language Processing (NLP). However, existing Mamba-based MIL approaches face three critical challenges: 
(1) disruption of 2D spatial locality during 1D sequence flattening; 
(2) sub-optimal modeling of fine-grained local cellular structures; and 
(3) high memory peaks during inference on resource-constrained edge devices. 
Studies like MambaOut reveal that Mamba's SSM component is redundant for local feature extraction, where Gated CNNs suffice. Recognizing that WSI analysis demands both fine-grained local feature extraction akin to natural images, and global context modeling akin to NLP, we propose MambaBack, a novel hybrid architecture that harmonizes the strengths of Mamba and MambaOut. 
First, we propose the Hilbert sampling strategy to preserve the 2D spatial locality of tiles within 1D sequences, enhancing the model's spatial perception. 
Second, we design a hierarchical structure comprising a 1D Gated CNN block based on MambaOut to capture local cellular features, and a BiMamba2 block to aggregate global context, jointly enhancing multi-scale representation. 
Finally, we implement an asymmetric chunking design, allowing parallel processing during training and chunking-streaming accumulation during inference, minimizing peak memory usage for deployment. 
Experimental results on five datasets demonstrate that MambaBack outperforms seven state-of-the-art methods. Source code and datasets are publicly available.
\end{abstract}

\vspace{0.35cm}
\section{Introduction}
Whole Slide Image (WSI) analysis is fundamental to modern pathology practice, playing a pivotal role in accurate cancer diagnosis~\cite{digital,review}. While WSIs preserve multi-scale morphological features ranging from the cellular to the tissue level, their gigapixel scale presents a significant challenge. Navigating these enormous images to identify critical, disease-relevant regions is labor-intensive and error-prone, thereby aggravating the diagnostic burden on pathologists~\cite{survey}.

With the advancement of deep learning, computational pathology has emerged as a powerful auxiliary tool for cancer diagnosis~\cite{deep}. Consequently, WSI analysis has evolved from computationally expensive end-to-end approaches to the more scalable Multiple Instance Learning (MIL) paradigm~\cite{clinical,abmil}. In current two-stage MIL, a frozen foundation model (e.g., UNI~\cite{uni}) first extracts dense feature embeddings from small tiles (e.g., 224$\times$224 pixels) partitioned from a WSI, followed by a learnable MIL aggregator processes these embeddings to generate slide-level predictions. By decoupling tile-level feature extraction from slide-level modeling, MIL enables a workflow that is both efficient and effective.

Early MIL methods are based on attention mechanism (e.g., ABMIL~\cite{abmil}, CLAM~\cite{clam}, DTFD-MIL~\cite{dtfd}). Although effective, these permutation-invariant approaches fail to capture complex inter-cellular relationships and spatial dependencies, limiting their ability to model comprehensive tissue morphology. To address this, Transformer-based methods were introduced to enhance morphological modeling via permutation-variant mechanisms (e.g., TransMIL~\cite{transmil}). However, despite their success on large-scale datasets, Transformers are prone to overfitting on smaller cohorts, a common scenario in medical imaging due to the high cost of annotation~\cite{cpia}. Recently, State Space Models (SSMs) like Mamba~\cite{mamba} have surpassed Transformers in MIL tasks (e.g., MambaMIL~\cite{mambamil}), offering linear complexity while retaining the benefits of permutation-variant modeling. Despite these advances, three critical challenges remain unresolved.

First, the intrinsic 2D spatial relationships of tissue are often disrupted when WSIs are flattened into 1D tile sequences~\cite{transmil}. Existing strategies like sequence reordering~\cite{mambamil} or Z-order~\cite{mambawsi} remain suboptimal for the non-convex, irregular shapes characteristic of pathological tissue. To respect the topological continuity of biological structures, we propose a Hilbert sampling strategy. By utilizing the space-filling properties of the Hilbert curve~\cite{hilbert}, we maximize the preservation of local spatial neighborhoods within the linearized sequence, ensuring that the model maintains a coherent perception of the tissue micro-environment.

Second, WSI analysis requires a delicate balance between extracting fine-grained cellular details and aggregating slide-level context~\cite{cellmix}. While SSMs excel at efficient long-context modeling, they are often less effective than Convolutional Neural Networks (CNNs) at extracting the subtle, homogeneous patterns of local cell structures. We therefore propose a hierarchical hybrid structure that mirrors the pathologist's zoom logic: a local 1D Gated CNN block based on the 2D Gated CNN block from MambaOut~\cite{mambaout} captures high-magnification cellular features, while a global BiMamba2 block based on BiMamba~\cite{vision_mamba} and Mamba2~\cite{mamba2} aggregates these into a comprehensive tissue-level representation.

\begin{figure}[t]
    \centering
    \includegraphics[width=1\linewidth]{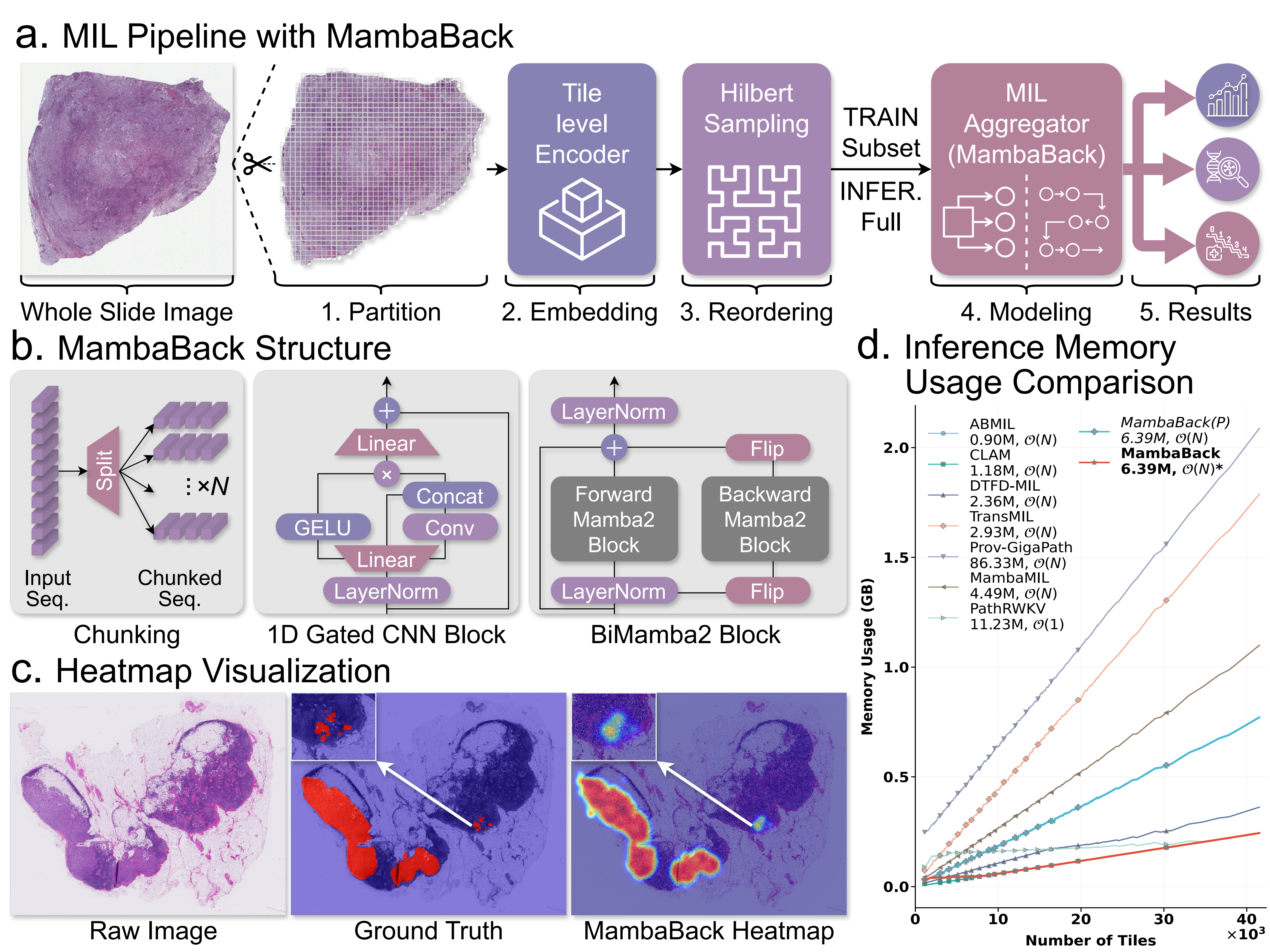}
    \caption{Overview of MambaBack. \textbf{a.} MIL pipeline with MambaBack. \textbf{b.} MambaBack structure. \textbf{c.} Heatmap visualization. \textbf{d.} Inference memory usage comparison.}
    \label{fig:main}
\end{figure}

Finally, high memory peaks during inference pose a significant barrier to deployment on resource-constrained edge devices~\cite{edge}. While MIL models are typically trained on high-performance GPUs, clinical deployment often targets hardware with limited memory (e.g., FPGAs)~\cite{fpga} to reduce cost. Ensuring memory efficiency and stability is therefore crucial, especially given the variable and massive dimensions of WSIs. Although prior work like PathRWKV~\cite{pathrwkv} achieved constant spatial complexity via an asymmetric design, it relies on max pooling aggregation. This has been shown to be suboptimal for complex prognostication tasks like survival analysis compared to attention-based pooling~\cite{deep_mil}. We effectively tackle this by designing an asymmetric chunking strategy that realizes constant-like memory complexity while preserving effective attention-based aggregation for better generalizability. In summary, MambaBack establishes a spatially-aware framework designed to harmonize micro-scale feature extraction with macro-scale context modeling. Our contributions are as follows:
\begin{itemize}
\item We propose a hybrid architecture that combines MambaOut with Mamba to harmonize fine-grained feature extraction and global context modeling.
\item We propose a Hilbert sampling strategy to maximally preserve 2D spatial locality within flattened sequences, and an asymmetric chunking mechanism that enables memory-efficient inference on resource-constrained devices.
\item Extensive experiments on 17,828 WSIs demonstrate that MambaBack outperforms seven state-of-the-art methods across five public datasets, validating its potential for robust computational pathology.
\end{itemize}

\section{Methods}
\subsection{MIL Pipeline with MambaBack}
As shown in Fig.~\ref{fig:main}a, WSIs are processed via standard non-overlapping tiling and quality control protocols~\cite{unpuzzle} to remove non-informative regions, followed by feature extraction using a foundation model (e.g., Prov-GigaPath~\cite{gigapath}). The extracted tile features are reorganized via a Hilbert sampling strategy before being processed by the MIL aggregator, MambaBack.

During the training phase, we set the batch size $B>1$ for training efficiency and stable gradient descent, and input sequence lengths are fixed to satisfy batching requirements. During inference, we set $B=1$ for precise inference, and all tile features within a slide are processed in their Hilbert-sorted order.

MambaBack aggregates reordered tile sequence to generate slide-level predictions. Notably, during the training phase, the model processes the entire tile sequence simultaneously, leveraging GPU parallelization for maximum efficiency. During inference, the tile sequence are partitioned into equal-sized chunks on the CPU and sequentially fed into the model for optimal memory utilization.

\subsection{Hilbert Sampling Strategy}
To preserve the 2D spatial structure of WSIs within 1D sequence, we propose a Hilbert sampling strategy. Unlike row-major flattening that disrupts vertical correlations, the Hilbert space-filling curve maintains spatial locality by clustering 2D neighbors into 1D proximity~\cite{hilbert}. The process comprises three steps: dense coordinate mapping, recursive Hilbert sequencing, and contiguous chunking.

Since tissue regions are sparse, raw tile coordinates $\mathcal{C}_{raw} = \{(x_i, y_i)\}_{i=1}^{N}$ are often discontinuous. We first map them to a dense rank-based grid to eliminate empty background space:
\begin{equation}
\tilde{x}_i = \text{Rank}(x_i), \quad \tilde{y}_i = \text{Rank}(y_i)
\end{equation}
where $\text{Rank}(\cdot)$ computes the ordinal rank among unique coordinate values.

We then define the Hilbert mapping on a grid of order $k$ (where $2^k \ge \max(\tilde{x}, \tilde{y}$)). We compute the Hilbert index $h_i$ recursively via quadrant traversal. At each level $l \in \{k-1,\dots,0\}$, the coordinate ($\tilde{x}_i, \tilde{y}_i$) falls into one of four quadrants, determining the local traversal order $q_l \in \{0,1,2,3\}$. To ensure continuity, the coordinate system is rotated or reflected based on the curve's entrance and exit points in the previous level. The final index $h_i$ is the summation of these weighted quadrant positions:
\begin{equation}
    h_i = \sum_{l=0}^{k-1} 4^l \cdot \phi_l(\tilde{x}_i, \tilde{y}_i, \Omega_{l+1})
\end{equation}
where $\phi_l$ determines the local index based on the current coordinates and the cumulative rotation or reflection state $\Omega_{l+1}$ propagated from higher levels.

Tile featuress are then sorted by $h_i$ to form the sequence $S$. To accommodate fixed input sizes $M$ during training while preserving the local context modeled by Mamba, we employ a stochastic contiguous chunking strategy, selecting a continuous segment from the sorted sequence. If $N>M$, we sample a random start index $\delta \sim \mathcal{U}[0, N-M]$ to obtain the training sequence $S 
_{train}$:
\begin{equation}
\mathcal{S}_{train} = { t{\pi(j)} \mid j = \delta, \delta+1, \dots, \delta+M-1 }
\end{equation}
This strategy acts as a 1D spatial data augmentation, ensuring that the model learns from continuous, topology-preserving tissue segments.

\subsection{MambaBack Structure}
MambaBack is designed as a hierarchical dual-stage framework to effectively process the massive number of tile features in WSIs. It consists of two primary components: a 1D Gated CNN based on the 2D Gated CNN from MambaOut as the local feature extractor, and a bidirectional Mamba2 (BiMamba2) based on Mamba2 as the global context aggregator (Fig.~\ref{fig:main}b).

To tackle the sequence length challenge, the input tile sequence $X \in \mathbb{R}^{N \times D}$ is partitioned into local segments. We define a chunk size $L$, reshaping the sequence into $M = \lceil N / L \rceil$ local chunks. For each chunk $C_i \in \mathbb{R}^{L \times D}$, we employ the 1D Gated CNN Block to model local correlations:
\begin{equation}
    \begin{gathered}
        \hat{C}_i = \text{LayerNorm}(C_i) \quad X_{gate}, X_{pass}, X_{conv} = \text{Split}(\text{Linear}_{in}(\hat{C}_i)) \\
        Z_i = \text{Linear}_{out}(\sigma(X_{gate}) \odot \text{Concat}(X_{pass}, \text{DWConv}(X_{conv}))) + C_i
    \end{gathered}
\end{equation}
where $\sigma$ is the GELU activation, Split divides the features along the channel dimension, and DWConv denotes a depth-wise convolution that captures local neighborhood information. The processed chunk features are then compressed into a single representative token $t_i$ via a local Gated Attention block. This step reduces the sequence length from $N$ to $M$, significantly lowering the computational burden for the subsequent stage.

To model the global dependencies across the entire slide, we process the sequence of aggregated tokens $T=\{t_1,\dots,t_
M\}$ using the BiMamba block. Unlike standard unidirectional SSMs, BiMamba captures context from both forward and backward directions:
\begin{equation}
\begin{gathered}
\hat{T} = \text{LayerNorm}(T) \\
H_{fwd} = \text{Mamba2}(\hat{T}) \quad H_{bwd} = \text{Flip}(\text{Mamba2}(\text{Flip}(\hat{T}))) \\
T_{global} = \text{LayerNorm}(T + H_{fwd} + H_{bwd})
\end{gathered}
\end{equation}
Finally, a global Gated Attention block aggregates $T_{global}$ into the slide-level embedding for classification.

\subsection{Asymmetric Design}
The inherent variability in tissue dimensions leads to massive fluctuations in sequence length $N$, posing a significant risk of Out-Of-Memory (OOM) errors when deployed on memory constrained edge devices. We therefore propose an asymmetric design that decouples the memory requirements of the local feature extraction stage from the total slide size during inference, while maintaining high throughput during training.

During training, computational efficiency is prioritized. The input sequence $X \in \mathbb{R}^{N \times D}$ is reshaped into $M = \lceil N/L \rceil$ chunks of size $L$ (where $L=64$). To maximize GPU saturation, we process all $M$ chunks in parallel through the local 1D Gated CNN and Gated Attention:
\begin{equation}
    T_{local} = \text{GatedAttention}(\text{GatedCNN1D}(\text{Reshape}(X))) \in \mathbb{R}^{M \times D}
\end{equation}
By compressing the sequence length from $N$ to $M$ in a single parallel pass, the subsequent global BiMamba2 block operates on a highly compact representation, significantly reducing the computational complexity of modeling global context.

During inference, we adopt a Chunk-and-Accumulate design to clamp peak memory usage. Instead of processing the entire slide at once, the algorithm iterates through the sequence in mini-batches of size $B_{inf}$:
\begin{equation}
    \begin{aligned}
        & T_{batch}^{(k)} = \text{LocalStage}(X_{k \cdot B_{inf} : (k+1) \cdot B_{inf}}) \\
        & T_{global} = \text{Concat}([T_{batch}^{(0)}, \dots, T_{batch}^{(K)}])
    \end{aligned}
\end{equation}
Crucially, the heavy intermediate feature maps of each mini-batch are immediately discarded after extracting the representative tokens $T_{batch}^{(k)}$. Therefore, local-stage peak memory is constant in $N$. Finally, the aggregated global context $T_{global}$, which is reduced by a factor of $L$, allows the global BiMamba block to perform slide-level classification with negligible memory overhead.
\section{Experiments}
\subsection{Datasets, Tasks, and Implementation Details}
We conduct comprehensive experiments on 5 datasets covering 4 downstream tasks to evaluate model performance: \textbf{CAMELYON16}~\cite{cam16} for binary breast metastasis classification on lymph nodes; 
\textbf{CAMELYON17}~\cite{cam17} for multi-class breast metastasis classification; 
\textbf{PANDA}~\cite{panda} for ISUP grading to assess prostate cancer aggressiveness; 
\textbf{TCGA-NSCLC}~\cite{tcga} for cancer subtyping distinguishing between normal, lung adenocarcinoma, and lung squamous cell carcinoma tissues; 
and \textbf{TCGA-BRCA}~\cite{tcga} for overall survival prediction, estimating patient survival time in months based on breast tissue morphology.

\begin{figure}[t]
    \centering
    \includegraphics[width=1\linewidth]{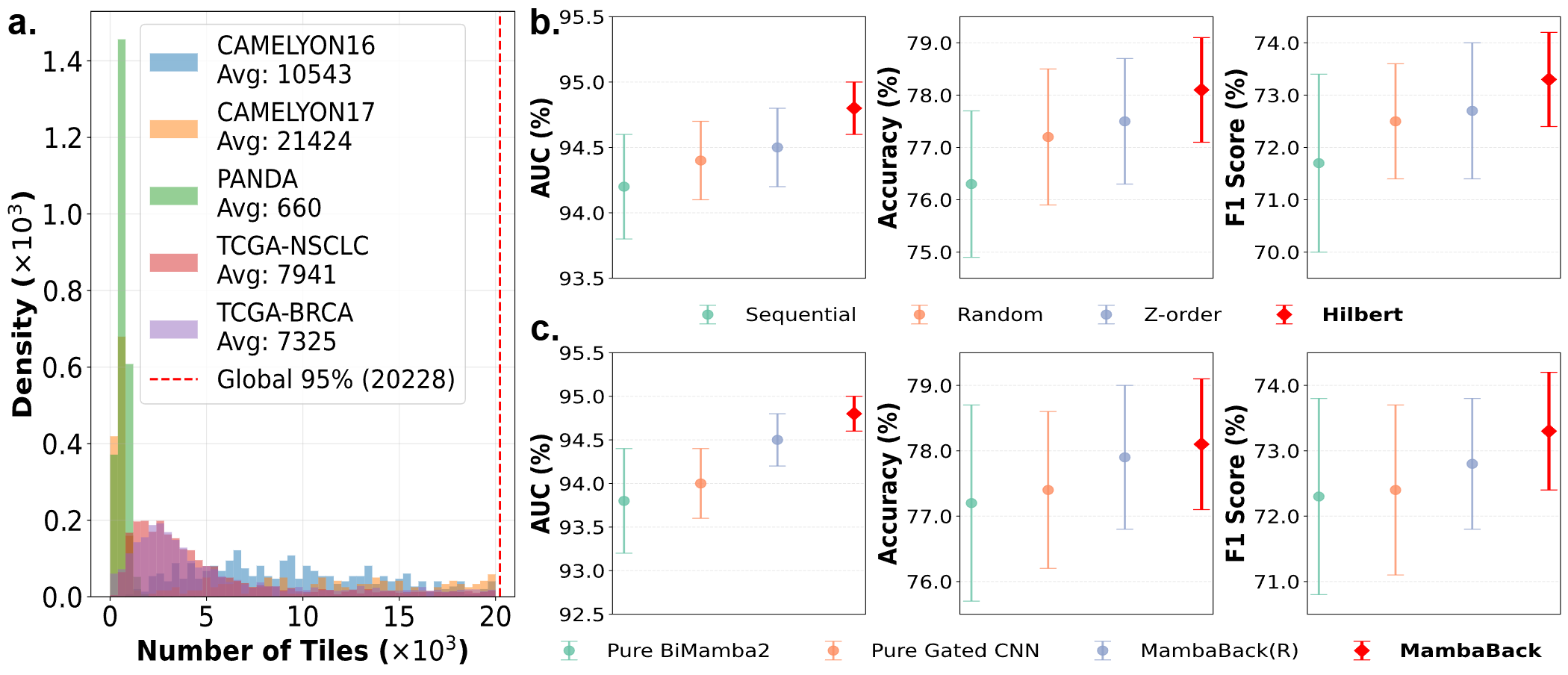}
    \caption{Analysis of WSI tile distributions and ablation studies on key components. \textbf{a.} Distribution of tile counts across datasets. \textbf{b.} Ablation study on sampling strategies on PANDA dataset. \textbf{c.} Ablation study on model architectures on PANDA dataset.}
    \label{fig:ablation}
\end{figure}

We implement our pipeline using the UnPuzzle framework~\cite{unpuzzle}. WSIs are partitioned into $224 \times 224$ tiles at 0.5 mpp and embedded via the frozen Prov-GigaPath encoder~\cite{gigapath}. Models are trained from scratch using AdamW optimizer and cosine decay scheduler. We optimize learning rates via grid search ($1\times10^{-4}$--$1\times10^{-2}$) and utilize early stopping with patience of 10 epochs. We utilize cross-entropy loss for classification and CoxPH loss~\cite{deepsurv} for survival tasks. We adopt Hilbert sampling strategy on all methods for both training and inference stages. During training, we employ a batch size $B=8$, maximum 4,096 tile features, except for PANDA, where we adjust to $B=64$ and maximum 512 tile features due to distribution variances (Fig.~\ref{fig:ablation}a). During inference, we utilize the best validation checkpoint to process all tile features with $B=1$. For metrics, we report C-Index for survival tasks, and Accuracy, AUC, and F1-Score for classification tasks, representing the average across 5 folds. Implementation relies on Python 3.12, PyTorch 2.9, and CUDA 12.8 on four NVIDIA RTX 4090 GPUs.

\begin{table}[t]
\caption{The performance comparison with SOTA methods.}
\label{tab:main}
\centering
\resizebox{\textwidth}{!}{
\begin{tabular}{c|c|cccccccc}
\toprule
\multirow{2}{*}{\diagbox{\textbf{Dataset}}{\textbf{Method}}} & \multirow{2}{*}{\textbf{Metric}} & \textbf{ABMIL} & \textbf{CLAM} & \textbf{DTFD-MIL} & \textbf{TransMIL} & \textbf{GigaPath} & \textbf{MambaMIL} & \textbf{PathRWKV} & \textbf{MambaBack} \\
& & \textbf{\cite{abmil}} & \textbf{\cite{clam}} & \textbf{\cite{dtfd}} & \textbf{\cite{transmil}} & \textbf{\cite{gigapath}} & \textbf{\cite{mambamil}} & \textbf{\cite{pathrwkv}} & \textbf{(Ours)} \\
\midrule
\textbf{CAMELYON16} & \textbf{AUC} & 0.989±0.005 & 0.990±0.003 & 0.990±0.006 & 0.993±0.001 & 0.987±0.007 & 0.992±0.003 & 0.991±0.001 & \textbf{0.995±0.004} \\
\textbf{Task: Cls. (2)} & \textbf{Acc.} & 0.980±0.000 & 0.981±0.004 & 0.981±0.004 & 0.982±0.004 & 0.961±0.024 & 0.983±0.008 & 0.983±0.008 & \textbf{0.984±0.008} \\
\textbf{Samples: 400} & \textbf{F1} & 0.978±0.000 & 0.980±0.005 & 0.980±0.005 & 0.981±0.005 & 0.958±0.026 & 0.982±0.009 & 0.981±0.008 & \textbf{0.983±0.008} \\
\midrule
\textbf{CAMELYON17} & \textbf{AUC} & 0.708±0.030 & 0.710±0.038 & 0.716±0.026 & 0.700±0.016 & 0.697±0.035 & 0.719±0.027 & 0.722±0.030 & \textbf{0.729±0.032} \\
\textbf{Task: Cls. (4)} & \textbf{Acc.} & 0.756±0.018 & 0.758±0.024 & 0.756±0.013 & 0.754±0.023 & 0.731±0.017 & 0.758±0.029 & 0.752±0.053 & \textbf{0.774±0.012} \\
\textbf{Samples: 500} & \textbf{F1} & 0.498±0.009 & 0.503±0.071 & 0.463±0.053 & 0.404±0.024 & 0.423±0.042 & 0.460±0.029 & 0.499±0.056 & \textbf{0.513±0.038} \\
\midrule
\textbf{PANDA} & \textbf{AUC} & 0.946±0.002 & 0.947±0.001 & 0.944±0.001 & 0.936±0.002 & 0.940±0.001 & 0.942±0.002 & 0.947±0.003 & \textbf{0.948±0.002} \\
\textbf{Task: Cls. (6)} & \textbf{Acc.} & 0.768±0.004 & 0.774±0.005 & 0.762±0.006 & 0.739±0.006 & 0.752±0.008 & 0.759±0.007 & 0.776±0.005 & \textbf{0.781±0.010} \\
\textbf{Samples: 10,616} & \textbf{F1} & 0.711±0.006 & 0.721±0.007 & 0.705±0.005 & 0.672±0.011 & 0.692±0.013 & 0.693±0.014 & 0.726±0.007 & \textbf{0.733±0.009} \\
\midrule
\textbf{TCGA-NSCLC} & \textbf{AUC} & 0.708±0.014 & 0.708±0.012 & 0.710±0.008 & 0.711±0.009 & 0.661±0.066 & 0.706±0.007 & 0.707±0.004 & \textbf{0.713±0.009} \\
\textbf{Task: Cls. (3)} & \textbf{Acc.} & 0.765±0.013 & 0.770±0.007 & 0.769±0.023 & 0.773±0.013 & 0.771±0.010 & 0.771±0.011 & 0.770±0.012 & \textbf{0.774±0.013} \\
\textbf{Samples: 3,210} & \textbf{F1} & 0.376±0.046 & 0.398±0.032 & 0.403±0.023 & 0.379±0.021 & 0.335±0.056 & 0.420±0.058 & 0.430±0.050 & \textbf{0.431±0.047} \\
\midrule
\textbf{TCGA-BRCA} & \multirow{3}{*}{\textbf{C-Index}} & \multirow{3}{*}{0.613±0.024} & \multirow{3}{*}{0.617±0.025} & \multirow{3}{*}{0.613±0.037} & \multirow{3}{*}{0.620±0.030} & \multirow{3}{*}{0.590±0.014} & \multirow{3}{*}{0.616±0.028} & \multirow{3}{*}{0.616±0.044} & \multirow{3}{*}{\textbf{0.623±0.033}} \\
\textbf{Task: Surv.} & & & & & & & & \\
\textbf{Samples: 3,102} & & & & & & & & \\
\bottomrule
\end{tabular}}
\end{table}

\subsection{Comparison Results}
We compared our work with seven SOTA methods: \textbf{ABMIL~\cite{abmil}}, \textbf{CLAM~\cite{clam}}, \textbf{DTFD-MIL~\cite{dtfd}}, \textbf{TransMIL~\cite{transmil}}, \textbf{Prov-GigaPath~\cite{gigapath}}, \textbf{MambaMIL~\cite{mambamil}}, and \textbf{PathRWKV~\cite{pathrwkv}} to comprehensively evaluate the robustness of our work. 

Tab.~\ref{tab:main} shows MambaBack achieves SOTA performance across all 5 benchmark datasets, consistently outperforming seven baseline methods across all metrics. Notably, SSMs (MambaMIL, PathRWKV, MambaBack) demonstrate superior stability and generalization capabilities compared to Transformers (TransMIL, Prov-GigaPath), which appear more prone to overfitting. Moreover, the substantial improvements in F1-score (e.g., 0.513 on CAMELYON17 and 0.733 on PANDA) validate the effectiveness of our hierarchical structure. Meanwhile, the heatmap visualization in Fig.~\ref{fig:main}c exhibits high concordance between ground truth annotations and the MambaBack attention maps, particularly in small regions. Collectively, these results demonstrate that by leveraging Gated CNNs for fine-grained local feature extraction, MambaBack captures subtle morphological details, ensuring a balance between precision and recall.

\subsection{Ablation Studies}
\subsubsection{Hilbert Sampling Strategy.}
We compared our proposed Hilbert sampling strategy against sequential sampling, random sampling, and Z-order sampling. Fig.~\ref{fig:ablation}b shows that Hilbert sampling consistently achieves superior performance with the lowest variance across all metrics on the PANDA dataset, validating its ability to preserve 2D spatial locality within 1D sequences.

\subsubsection{Hybrid Structure.}
We compared MambaBack against pure BiMamba2, pure Gated CNN, and MambaBack(R) (local BiMamba2 with global Gated CNN). Fig.~\ref{fig:ablation}c shows that our structure consistently outperforms others across all metrics on the PANDA dataset, confirming the necessity of our hybrid design in harmonizing fine-grained local features with global context.

\subsubsection{Asymmetric Design.}
We compared inference memory usage as shown in Fig.~\ref{fig:main}d. Unlike Transformers (Prov-GigaPath, TransMIL) that exhibit high peak memory, or PathRWKV that maintains a high baseline for typical slide counts ($<40,000$), MambaBack(P) (MambaBack with parallel training structure) achieves lower memory usage compared to MambaMIL, owing to the Gated CNN. Furthermore, MambaBack mirrors the low memory profile of lightweight models like CLAM. This demonstrates that our design achieves deployment costs comparable to lightweight models based on a more comprehensive architecture.
\section{Conclusion}
In this work, we present MambaBack, a hybrid model with Hilbert sampling strategy. Experiments on five diverse datasets and tasks against seven SOTA methods demonstrate that MambaBack exhibits superior performance and generalizability. Crucially, by leveraging an asymmetric design that decouples training throughput from inference memory costs, MambaBack breaks the computational barriers of deployment, offering a scalable and efficient solution for real-time pathological assessment on resource-constrained clinical edge devices.
\bibliography{Sections/Reference}

@article{digital,
  title={Digital pathology and artificial intelligence},
  author={Niazi, Muhammad Khalid Khan and Parwani, Anil V and Gurcan, Metin N},
  journal={The lancet oncology},
  volume={20},
  number={5},
  pages={e253--e261},
  year={2019},
  publisher={Elsevier}
}

@article{review,
  title={Review of the current state of whole slide imaging in pathology},
  author={Pantanowitz, Liron and Valenstein, Paul N and Evans, Andrew J and Kaplan, Keith J and Pfeifer, John D and Wilbur, David C and Collins, Laura C and Colgan, Terence J},
  journal={Journal of pathology informatics},
  volume={2},
  number={1},
  pages={36},
  year={2011},
  publisher={Elsevier}
}

@article{survey,
  title={A survey on deep learning in medical image analysis},
  author={Litjens, Geert and Kooi, Thijs and Bejnordi, Babak Ehteshami and Setio, Arnaud Arindra Adiyoso and Ciompi, Francesco and Ghafoorian, Mohsen and Van Der Laak, Jeroen Awm and Van Ginneken, Bram and S{\'a}nchez, Clara I},
  journal={Medical image analysis},
  volume={42},
  pages={60--88},
  year={2017},
  publisher={Elsevier}
}

@article{deep,
  title={Deep neural network models for computational histopathology: A survey},
  author={Srinidhi, Chetan L and Ciga, Ozan and Martel, Anne L},
  journal={Medical image analysis},
  volume={67},
  pages={101813},
  year={2021},
  publisher={Elsevier}
}

@inproceedings{abmil,
  title={Attention-based deep multiple instance learning},
  author={Ilse, Maximilian and Tomczak, Jakub and Welling, Max},
  booktitle={International conference on machine learning},
  pages={2127--2136},
  year={2018},
  organization={PMLR}
}

@article{clinical,
  title={Clinical-grade computational pathology using weakly supervised deep learning on whole slide images},
  author={Campanella, Gabriele and Hanna, Matthew G and Geneslaw, Luke and Miraflor, Allen and Werneck Krauss Silva, Vitor and Busam, Klaus J and Brogi, Edi and Reuter, Victor E and Klimstra, David S and Fuchs, Thomas J},
  journal={Nature medicine},
  volume={25},
  number={8},
  pages={1301--1309},
  year={2019},
  publisher={Nature Publishing Group US New York}
}

@article{uni,
  title={Towards a general-purpose foundation model for computational pathology},
  author={Chen, Richard J and Ding, Tong and Lu, Ming Y and Williamson, Drew FK and Jaume, Guillaume and Song, Andrew H and Chen, Bowen and Zhang, Andrew and Shao, Daniel and Shaban, Muhammad and others},
  journal={Nature medicine},
  volume={30},
  number={3},
  pages={850--862},
  year={2024},
  publisher={Nature Publishing Group US New York}
}

@article{gigapath,
  title={A whole-slide foundation model for digital pathology from real-world data},
  author={Xu, Hanwen and Usuyama, Naoto and Bagga, Jaspreet and Zhang, Sheng and Rao, Rajesh and Naumann, Tristan and Wong, Cliff and Gero, Zelalem and Gonz{\'a}lez, Javier and Gu, Yu and others},
  journal={Nature},
  pages={1--8},
  year={2024},
  publisher={Nature Publishing Group UK London}
}

@article{clam,
  title={Data-efficient and weakly supervised computational pathology on whole-slide images},
  author={Lu, Ming Y and Williamson, Drew FK and Chen, Tiffany Y and Chen, Richard J and Barbieri, Matteo and Mahmood, Faisal},
  journal={Nature biomedical engineering},
  volume={5},
  number={6},
  pages={555--570},
  year={2021},
  publisher={Nature Publishing Group UK London}
}

@inproceedings{dtfd,
  title={Dtfd-mil: Double-tier feature distillation multiple instance learning for histopathology whole slide image classification},
  author={Zhang, Hongrun and Meng, Yanda and Zhao, Yitian and Qiao, Yihong and Yang, Xiaoyun and Coupland, Sarah E and Zheng, Yalin},
  booktitle={Proceedings of the IEEE/CVF conference on computer vision and pattern recognition},
  pages={18802--18812},
  year={2022}
}

@article{transmil,
  title={Transmil: Transformer based correlated multiple instance learning for whole slide image classification},
  author={Shao, Zhuchen and Bian, Hao and Chen, Yang and Wang, Yifeng and Zhang, Jian and Ji, Xiangyang and others},
  journal={Advances in neural information processing systems},
  volume={34},
  pages={2136--2147},
  year={2021}
}

@article{cpia,
  title={CPIA dataset: a large-scale comprehensive pathological image analysis dataset for self-supervised learning pre-training},
  author={Ying, Nan and Lei, Yanli and Zhang, Tianyi and Lyu, Shangqing and Chen, Sicheng and Liu, Zeyu and Feng, Yunlu and Zhao, Yu and Zhang, Guanglei},
  journal={Biomedical Signal Processing and Control},
  volume={110},
  pages={108148},
  year={2025},
  publisher={Elsevier}
}

@inproceedings{mamba,
  title={Mamba: Linear-time sequence modeling with selective state spaces},
  author={Gu, Albert and Dao, Tri},
  booktitle={First conference on language modeling},
  year={2024}
}

@inproceedings{mambamil,
  title={Mambamil: Enhancing long sequence modeling with sequence reordering in computational pathology},
  author={Yang, Shu and Wang, Yihui and Chen, Hao},
  booktitle={International conference on medical image computing and computer-assisted intervention},
  pages={296--306},
  year={2024},
  organization={Springer}
}

@article{mambawsi,
  title={Exploring Multi-Scale Local and Global Features in Whole Slide Images Using State Space Models},
  author={Jiang, Chongcong and Zhao, Zhuo and Liang, Peixian and Shi, Min and Han, Jun and Tzeng, Nian-Feng and Xiao, Guanghua and Chen, Danny Z and Zheng, Hao},
  journal={bioRxiv},
  pages={2026--01},
  year={2026},
  publisher={Cold Spring Harbor Laboratory}
}

@book{hilbert,
  title={Space-filling curves},
  author={Sagan, Hans},
  year={2012},
  publisher={Springer Science \& Business Media}
}

@article{cellmix,
  title={CellMix: A General Instance Relationship-Based Method for Data Augmentation Toward Pathology Image Classification},
  author={Zhang, Tianyi and Yan, Zhiling and Li, Chunhui and Ying, Nan and Lei, Yanli and Lyu, Shangqing and Feng, Yunlu and Zhao, Yu and Zhang, Guanglei},
  journal={IEEE Transactions on Neural Networks and Learning Systems},
  year={2025},
  publisher={IEEE}
}

@inproceedings{mambaout,
  title={Mambaout: Do we really need mamba for vision?},
  author={Yu, Weihao and Wang, Xinchao},
  booktitle={Proceedings of the Computer Vision and Pattern Recognition Conference},
  pages={4484--4496},
  year={2025}
}

@article{vision_mamba,
  title={Vision mamba: Efficient visual representation learning with bidirectional state space model},
  author={Zhu, Lianghui and Liao, Bencheng and Zhang, Qian and Wang, Xinlong and Liu, Wenyu and Wang, Xinggang},
  journal={arXiv preprint arXiv:2401.09417},
  year={2024}
}

@article{mamba2,
  title={Transformers are ssms: Generalized models and efficient algorithms through structured state space duality},
  author={Dao, Tri and Gu, Albert},
  journal={arXiv preprint arXiv:2405.21060},
  year={2024}
}

@article{edge,
  title={Deep learning with edge computing: A review},
  author={Chen, Jiasi and Ran, Xukan},
  journal={Proceedings of the IEEE},
  volume={107},
  number={8},
  pages={1655--1674},
  year={2019},
  publisher={IEEE}
}

@article{fpga,
  title={A survey of FPGA-based neural network accelerator},
  author={Guo, Kaiyuan and Zeng, Shulin and Yu, Jincheng and Wang, Yu and Yang, Huazhong},
  journal={arXiv preprint arXiv:1712.08934},
  year={2017}
}

@article{pathrwkv,
  title={PathRWKV: Enabling Whole Slide Prediction with Recurrent-Transformer},
  author={Chen, Sicheng and Zhang, Tianyi and Liao, Dankai and Li, Dandan and Han, Low Chang and Jiang, Yanqin and Jin, Yueming and Lyu, Shangqing},
  journal={arXiv preprint arXiv:2503.03199},
  year={2025}
}

@inproceedings{deep_mil,
  title={Deep multi-instance learning for survival prediction from whole slide images},
  author={Yao, Jiawen and Zhu, Xinliang and Huang, Junzhou},
  booktitle={International Conference on Medical Image Computing and Computer-Assisted Intervention},
  pages={496--504},
  year={2019},
  organization={Springer}
}

@article{unpuzzle,
  title={UnPuzzle: A Unified Framework for Pathology Image Analysis},
  author={Liao, Dankai and Chen, Sicheng and Xi, Nuwa and Xue, Qiaochu and Li, Jieyu and Hou, Lingxuan and Liu, Zeyu and Low, Chang Han and Wu, Yufeng and Liu, Yiling and others},
  journal={arXiv preprint arXiv:2503.03152},
  year={2025}
}

@article{panda,
  title={Artificial intelligence for diagnosis and Gleason grading of prostate cancer: the PANDA challenge},
  author={Bulten, Wouter and Kartasalo, Kimmo and Chen, Po-Hsuan Cameron and Str{\"o}m, Peter and Pinckaers, Hans and Nagpal, Kunal and Cai, Yuannan and Steiner, David F and Van Boven, Hester and Vink, Robert and others},
  journal={Nature medicine},
  volume={28},
  number={1},
  pages={154--163},
  year={2022},
  publisher={Nature Publishing Group US New York}
}

@article{cam16,
  title={Diagnostic assessment of deep learning algorithms for detection of lymph node metastases in women with breast cancer},
  author={Bejnordi, Babak Ehteshami and Veta, Mitko and Van Diest, Paul Johannes and Van Ginneken, Bram and Karssemeijer, Nico and Litjens, Geert and Van Der Laak, Jeroen AWM and Hermsen, Meyke and Manson, Quirine F and Balkenhol, Maschenka and others},
  journal={Jama},
  volume={318},
  number={22},
  pages={2199--2210},
  year={2017},
  publisher={American Medical Association}
}

@article{cam17,
  title={From detection of individual metastases to classification of lymph node status at the patient level: the camelyon17 challenge},
  author={Bandi, Peter and Geessink, Oscar and Manson, Quirine and Van Dijk, Marcory and Balkenhol, Maschenka and Hermsen, Meyke and Bejnordi, Babak Ehteshami and Lee, Byungjae and Paeng, Kyunghyun and Zhong, Aoxiao and others},
  journal={IEEE transactions on medical imaging},
  volume={38},
  number={2},
  pages={550--560},
  year={2018},
  publisher={IEEE}
}

@article{tcga,
  title={TCGAbiolinks: an R/Bioconductor package for integrative analysis of TCGA data},
  author={Colaprico, Antonio and Silva, Tiago C and Olsen, Catharina and Garofano, Luciano and Cava, Claudia and Garolini, Davide and Sabedot, Thais S and Malta, Tathiane M and Pagnotta, Stefano M and Castiglioni, Isabella and others},
  journal={Nucleic acids research},
  volume={44},
  number={8},
  pages={e71--e71},
  year={2016},
  publisher={Oxford University Press}
}

@article{deepsurv,
  title={DeepSurv: personalized treatment recommender system using a Cox proportional hazards deep neural network},
  author={Katzman, Jared L and Shaham, Uri and Cloninger, Alexander and Bates, Jonathan and Jiang, Tingting and Kluger, Yuval},
  journal={BMC medical research methodology},
  volume={18},
  number={1},
  pages={24},
  year={2018},
  publisher={Springer}
}
\end{document}